\documentclass{IEEEtran}
\pdfoutput=1

\usepackage{flushend}

\usepackage[utf8]{inputenc}
\usepackage[T1]{fontenc}

\usepackage{amssymb}
\usepackage{amsmath}

\newcommand{\XSenc}{\ensuremath{X_S^{\text{enc}}}}
\newcommand{\XSdec}{\ensuremath{X_S^{\text{dec}}}}

\newcommand{\XOenc}{\ensuremath{X_0^{\text{enc}}}}
\newcommand{\XOdec}{\ensuremath{X_0^{\text{dec}}}}

\newcommand{\XRenc}{\ensuremath{X_R^{\text{enc}}}}
\newcommand{\XRdec}{\ensuremath{X_R^{\text{dec}}}}

\usepackage{paralist}

\usepackage[numbers]{natbib}

\usepackage{tikz}
\usetikzlibrary{arrows,positioning}
\tikzset{blob/.style={circle,draw,font=\sffamily\bfseries,minimum size=3.5em,text centered}}
\tikzset{g/.style={gray!35}}
\tikzset{t/.style={semithick}}
\tikzset{dt/.style={t,dashed}}

\newcommand\indep{\protect\mathpalette{\protect\independenT}{\perp}}
\def\independenT#1#2{\mathrel{\rlap{$#1#2$}\mkern2mu{#1#2}}}

\usepackage[hidelinks]{hyperref}

\newcommand\copyrighttext{\footnotesize
The final version is published in \textit{Pattern Recognition in Neuroimaging, 2014 International Workshop on,} 1--4, 2014, \href{http://dx.doi.org/10.1109/PRNI.2014.6858551}{doi: 10.1109/PRNI.2014.6858551}.\\
\copyright\ 2014 IEEE. Personal use of this material is permitted. Permission from IEEE must be obtained for all other uses, in any current or future media, including reprinting/republishing this material for advertising or promotional purposes, creating new collective works, for resale or redistribution to servers or lists, or reuse of any copyrighted component of this work in other works.
}

\newcommand\copyrightnotice{%
\begin{tikzpicture}[remember picture,overlay]
\node[anchor=south,yshift=10pt] at (current page.south) {{\parbox{\dimexpr\textwidth-\fboxsep-\fboxrule\relax}{\copyrighttext}}};
\end{tikzpicture}%
}

\usepackage{textpos}

\begin{document}

\title{Causal and anti-causal learning in pattern recognition for neuroimaging}

\author{{ Sebastian Weichwald$^1$, Bernhard Schölkopf$^1$, Tonio Ball$^2$, Moritz Grosse-Wentrup$^1$\\\ \\
$^1$ Max Planck Institute for Intelligent Systems, Tübingen, Germany\\ {\small\texttt{\{sweichwald, bs, moritzgw\}@tuebingen.mpg.de}}\\
$^2$ Bernstein Center Freiburg, University of Freiburg, Freiburg, Germany\\ {\small\texttt{tonio.ball@uniklinik-freiburg.de}}}}

\maketitle
\copyrightnotice

\begin{abstract}
Pattern recognition in neuroimaging distinguishes between two types of models: encoding- and decoding models. This distinction is based on the insight that brain state features, that are found to be relevant in an experimental paradigm, carry a different meaning in encoding- than in decoding models. In this paper, we argue that this distinction is not sufficient: Relevant features in encoding- and decoding models carry a different meaning depending on whether they represent causal- or anti-causal relations. We provide a theoretical justification for this argument and conclude that causal inference is essential for interpretation in neuroimaging.\end{abstract}

\section{Introduction}

Pattern recognition in neuroimaging aims to provide insights into the neural basis of cognitive processes. Two types of models are used in this endeavor: encoding- and decoding models. Encoding models predict a subject's brain state for a given experimental condition, while decoding models aim to reconstruct experimental conditions from neuroimaging data. This difference has important consequences for the interpretation of brain state features that are found to be relevant in each type of model.

It has been argued that only encoding models can provide a complete functional description of a region of interest \cite{Naselaris2011}. Decoding models, on the other hand, may determine brain state features as relevant that are statistically independent of the experimental condition \cite{Todd2013}.
While in linear decoding models potential misinterpretations can be avoided by converting them into encoding models \cite{Haufe2014}, this is a substantially more difficult problem for non-linear decoding models. As decoding models are becoming ever more popular in the analysis of neuroimaging data \cite{Pereira2009}, the correct interpretation of such models is of considerable importance.

In this paper, we argue that the distinction between encoding- and decoding models is not sufficient to determine the meaning of relevant features in each type of model: Pattern recognition models need to be further distinguished with respect to whether they learn causal- or anti-causal relations \cite{Schoelkopf2012}.
In general, neuroimaging studies are based on the following causal structure: stimulus $\rightarrow$ brain activity $\rightarrow$ response.
We note that more complex experimental paradigms, in which responses again act as stimuli \cite{Gomez2011}, can also be modeled in this way by considering time-resolved variables, e.\,g.~stimulus$[t_0]$ $\rightarrow$ brain activity$[t_1]$ $\rightarrow$ response$[t_2]$.
Depending on whether experimental conditions are chosen to represent stimuli or responses, encoding- and decoding models then model causal- or anti-causal relations. In the following, we argue that this has important consequences for the interpretation of relevant features in each type of model. Furthermore, we argue that interpretation of neuroimaging data de facto requires causal inference problems to be solved.

The remainder of this article is organized as follows. In section~\ref{sec:methods} we introduce the necessary notation and terminology to formulate our proposed distinction of pattern recognition models in section~\ref{sec:propdist}. Next, we theoretically investigate the interpretability of relevant features in each type of pattern recognition model (sections \ref{sec:intA} to \ref{sec:intD}) and briefly summarize our findings in section \ref{sec:intE}. In section \ref{sec:causalinference} we argue that interpreting encoding- and decoding models is only a first step towards solving causal inference problems in the interpretation of neuroimaging data. We close with a conclusion in section~\ref{sec:conclusion}.\section{Pattern recognition models}\label{sec:methods}

\subsection{Notation}
By $X$ we denote the brain states represented by $d$ features obtained from neuroimaging data, i.\,e. $X = \{X_1,...,X_d\}$; by $Y$ we denote the (usually discrete) experimental conditions. Throughout this paper we use the notations $p(X)$, $p(X|Y)$ and $p(X,Y)$ for (conditional or joint) probability density functions (PDFs). All PDFs are assumed to be known.

Independence is denoted by $X\indep Y$ and conditional independence by $X \indep Y | Z$. Dependence and conditional dependence is denoted by $X \not\indep Y$ and $X \not \indep Y|Z$, respectively. Causal relations in a directed acyclic graph are denoted by $X \to Y$ \cite{Pearl2000}.

\subsection{Encoding and decoding models}
An encoding model $p(X|Y)$ represents how various experimental conditions are encoded in different brain states. We ask ``How does the brain state look like given a certain experimental condition?''. Examples for encoding models are the general linear model \cite{Friston1994} or the class-conditional mean: $E\{X|Y\}$.

A decoding model $p(Y|X)$ represents how different experimental conditions can be inferred from different brain states
\cite{Mitchell2004}. We ask ``Which experimental condition is most likely given a certain brain state?''.
Decoding models are for example obtained using support vector machines or linear regression.

Note that this distinction solely reflects the direction of modeling according to the brain state but neglects any causal relation between brain state and experimental condition that might be known a priori.

\subsection{Causal and anti-causal learning}
The brain is constantly exposed to the world's stimuli and processes them, e.\,g. giving raise to perceptions. As such, stimuli $S$ can only be causes but not effects of brain states $X$. The brain also constantly generates responses, e.\,g. movements, that are caused by the brain states. This gives rise to the following causal structure in neuroimaging studies: stimulus $\rightarrow$ brain state $\rightarrow$ response.
Note that we are not necessarily able to observe all stimuli that cause a certain brain state or all features of the brain state which are causal for $R$.
The causal structure enables us to distinguish between the following two scenarios:

\subsubsection{Stimulus-based experiments}
In a stimulus-based experiment the experimental conditions $Y$ correspond to stimuli $S$ presented to the subject. In general, we can control the stimulus presentation procedure and are thus able to randomize the presentation of stimuli. An example of a stimulus-based experiment is the randomized presentation of auditory stimuli to either the left or right ear. The causal structure of this setup is given by $S \rightarrow X$, i.\,e.~stimuli cause brain activity.

In this case the encoding model $p(X|Y) = p(X|S)$ represents a causal relation, while the decoding model $p(Y|X) = p(S|X)$ models an anti-causal relation.

\subsubsection{Response-based experiments}
In a response-based experiment the experimental conditions $Y$ represent subjects' responses that we observe. An example of a response-based experiment is the recording of volitional movements of either the left or right hand. The causal structure of this setup is given by $X \rightarrow R$, i.\,e.~brain activity causes responses. Note that in this setting we are not able to control for and randomize the experimental conditions.

In contrast to a stimulus-based experiment, the encoding model $p(X|Y) = p(X|R)$ of a response-based experiment represents an anti-causal relation, while the decoding model $p(Y|X) = p(R|X)$ models a causal relation.

\subsection{Distinction of pattern recognition models}\label{sec:propdist}
Considering both the distinction of encoding- and decoding models and the distinction of stimulus- and response-based experiments we obtain the following four types of models:

\begin{enumerate}[A.]
 \item Causal encoding models -- $p(X|S)$
 \item Anti-causal decoding models -- $p(S|X)$
 \item Anti-causal encoding models -- $p(X|R)$
 \item Causal decoding models -- $p(R|X)$
\end{enumerate}

In the following section we provide theoretical justifications why this distinction needs to be considered before interpreting encoding- or decoding models. As we show, interpretability of relevant features depends on whether the model represents causal or anti-causal relations.\section{Interpretation of relevant features}

When interpreting an encoding model $p(X|Y)$, we want to link features relevant for encoding to the experimental condition. Relevant here means that we determine the set of brain state features that the encoding model deems dependent on the experimental condition, i.\,e.~the features $X_i$ for which $p(X_i|Y) \neq p(X_i)$ and hence $X_i \not\indep Y$. The remaining features are independent of $Y$.
One way to do this in practice is to test the class-conditional sample means of each feature for statistically significant differences.
Features that, according to this univariate test, significantly vary with $Y$ are considered relevant for the encoding model.

When interpreting a decoding model $p(Y|X)$, we want to determine which features are relevant for decoding the experimental condition.
Relevant here means that we determine if a brain state feature or a set of features $X_i$ helps in decoding the experimental condition, i.\,e. it is tested whether $p(Y|X) \neq p(Y|X \setminus X_i)$ and hence $X_i \not\indep Y | X\setminus X_i$.
One way to do this in practice is recursive feature elimination, i.\,e. permuting or removing $X_i$ from the feature set and testing whether this significantly decreases decoding accuracy \cite{DeMartino2008}.
It is common to remove all features that are irrelevant for decoding to reduce dimensionality and obtain the minimal set of features that yields an optimal decoding model.
Features of that set are considered relevant for the decoding model.
We note that there might be other ways of identifying relevant features of a decoding model which might lead to different conclusions.

For our theoretical arguments we assume that we can identify all relevant features for each type of model.
We now show that relevant features in encoding- and decoding models carry a different meaning depending on the causal structure.

\subsection{Causal encoding models}\label{sec:intA}
From the encoding model $p(X|Y) = p(X|S)$ of a stimulus-based experiment we obtain the set $\XSenc$ of features that are dependent on $S$, i.\,e. for every $X_i \in \XSenc$ we have $S \not\indep X_i$. We denote the complementary set as $\XOenc := X \setminus \XSenc$.

According to Reichenbach's principle \cite{Reichenbach1956}, the dependency between $S$ and $X^{\text{enc}}_S$ implies that $S \rightarrow X^{\text{enc}}_S$, $S \leftarrow X^{\text{enc}}_S$, or $S \leftarrow H \rightarrow X^{\text{enc}}_S$ with $H$ a joint common cause of $S$ and $X^{\text{enc}}_S$. In the stimulus-based setting we can control for and randomize the stimulus. This enables us to rule out the last two cases and conclude $S \rightarrow X^{\text{enc}}_S$, i.\,e. the features in $X^{\text{enc}}_S$ are genuine effects of $S$ \cite{Holland1986}.

In addition, we have $S \indep \XOenc$, which allows us to conclude that features in $\XOenc$ are not genuine effects of $S$.

As such, all relevant features in a causal encoding model are genuine effects of $S$, while irrelevant features are not effects of $S$.

\subsection{Anti-causal decoding models}\label{sec:antidec}
From the decoding model $p(Y|X) = p(S|X)$ of a stimulus-based experiment we obtain the minimal set $\XSdec$ of features that allows to decode the stimulus, i.\,e. $p(S|X) = p(S|\XSdec)$. It hence holds that $S \indep \XOdec | X^\text{dec}_S$ where $\XOdec := X \setminus \XSdec$ is the set of features that do not further improve decoding.

We now describe two counterexamples that show that one can neither conclude that features in $\XOdec$ are not genuine effects of $S$ nor that features in $\XSdec$ are indeed genuine effects of $S$. First, assume $S \to X_1 \to X_2$. Since $p(S|X_1,X_2) = p(S|X_1)$, i.\,e. $S \indep X_2 | X_1$, we have $X_2 \in \XOdec$ although $X_2$ is actually a genuine effect of $S$. Second, assume $S \to X_1 \gets X_2$. Since $p(S|X_2,X_1) \neq p(S|X_1)$, i.\,e. $S \not\indep X_2|X_1$, we obtain $X_2 \in \XSdec$ although $X_2$ is not a genuine effect of $S$.

This establishes that interpreting anti-causal decoding models in this way has two drawbacks. First, features in $\XSdec$ can only be considered as potential effects of $S$. Second,  genuine effects of $S$ might be missed.

\subsection{Anti-causal encoding models}\label{sec:antienc}
Form the encoding model $p(X|Y) = p(X|R)$ of a response-based experiment we obtain the set of features that are dependent on $R$, i.\,e. for every $X_i \in \XRenc$ we have $X_i \not\indep R$. We denote the complementary set as $\XOenc := X \setminus \XRenc$ (overloading notation).

According to Reichenbach's principle, the dependency between $X_i \in X^{\text{enc}}_R$ and $R$ implies that $X_i \rightarrow R$, $X_i \leftarrow R$, or $X_i \leftarrow H \rightarrow R$ with $H$ a joint common cause of $X_i$ and $R$. A priori we know that brain activity $\to$ response. This enables us to rule out the case $X_i \gets R$. As we show next, we can not uniquely determine which of the last two scenarios is the case, i.\,e. features in $X^{\text{enc}}_R$ are potential but not necessarily genuine causes of $R$.

Consider $X_2 \gets X_1 \to R$: we have $X_1 \not\indep R$ and $X_2 \not\indep R$ and therefore $X_1,X_2 \in X_R^{\text{enc}}$. But note that $X_1 \to R$ while $X_2 \not \to R$, i.\,e. $X_2$ is not a cause of $R$. This shows that features in $\XRenc$ are not necessarily genuine causes of $R$.

Features in $\XOenc$, on the other hand, are independent of $R$ and can hence be considered to be no causes of $R$.

As such, not all relevant features in anti-causal encoding models are genuine causes of $R$, while irrelevant features are indeed not causal for $R$.

\subsection{Causal decoding models}\label{sec:intD}
From the decoding model $p(Y|X) = p(R|X)$ of a response-based experiment we obtain the minimal set $\XRdec$ of features that allows to decode the response, i.\,e. $p(R|X) = p(R|\XRdec)$. It hence holds that $S \indep \XOdec | X^\text{dec}_R$ where $\XOdec := X \setminus \XRdec$ is the set of features that do not further improve decoding.

\begin{figure}[t]
\centering
\begin{tikzpicture}[->,>=stealth,auto,on grid,ultra thick]
    \node[blob](a) at(0,2) {$X_1$};
    \node[blob](h) at(2,2) {$H$};
    \node[blob](b) at(4,2) {$X_2$};
    \node[blob](R) at(2,0) {$R$};

  \path[every node/.style={font=\sffamily\small}]
    (h) edge node [left] {} (a)
    (h) edge node [left] {} (b)
    (h) edge node [left] {} (R);
\end{tikzpicture}
\caption{Causal graph of an exemplary response-based experiment: $H$ is not observable as a brain state feature and hence denotes a hidden common cause of the observed brain state features $X_1$ and $X_2$ and the response $R$.}\label{fig:hidres}
\end{figure}
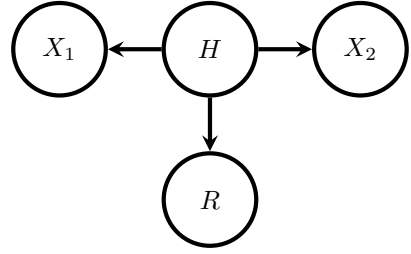

We now describe two counterexamples that show that one can neither conclude that features in $\XOdec$ are not genuine causes of $R$ nor that features in $\XRdec$ are genuine causes of $R$.
First, assume $X_2 \to X_1 \to R$. Since $p(R|X_1,X_2) = p(R|X_1)$, i.\,e. $X_2 \indep R | X_1$, we have $X_2 \in \XOdec$ although $X_2$ is a cause of $R$.
Second, assume the graph shown in figure~\ref{fig:hidres} where $H$ is a hidden common cause of $X_1,X_2$ and $R$ which is not observable as a brain state feature. Since $p(R|X_2,X_1) \neq p(R|X_1)$ and $p(R|X_2,X_1) \neq p(R|X_2)$ we have  $X_1,X_2 \in \XRdec$ although both $X_1$ and $X_2$ are not causes of $R$.

This establishes that interpreting causal decoding models this way has two drawbacks. First, features in $\XRdec$ are not necessarily causes of $R$. Second, genuine of $R$ causes might be missed.

\subsection{Subsumption}\label{sec:intE}
In the previous sections we showed that the interpretation of relevant features in encoding- and decoding models depends on the underlying causal structure. This justifies our argument that the distinction of encoding- and decoding models is not sufficient. In particular we argued that, without employing further assumptions,

\begin{enumerate}[A.]
        \item causal encoding models $p(X|S)$ allow to identify genuine effects $X_S$ of $S$.
        \item anti-causal decoding models $p(S|X)$ allow to identify some potential effects of $S$.
        \item anti-causal encoding models $p(X|R)$ allow to identify potential causes of $R$.
        \item causal decoding models $p(R|X)$ allow to identify some potential causes of $R$.
\end{enumerate}\section{Causal inference in neuroimaging}\label{sec:causalinference}

So far, we have argued that the causal structure of a neuroimaging study, i.\,e.~whether we learn in causal- or anti-causal direction, has to be taken into account when interpreting relevant features in encoding- and decoding models. In particular, we have shown that, with the exception of the causal encoding model, the meaning of relevant features in encoding- and decoding models is ambiguous. In the following, we demonstrate on two examples that such ambiguities can be resolved by means of causal inference \cite{Pearl2000,Spirtes2000}. Throughout this section we assume \textit{faithfulness}, i.\,e.~we assume that all observed (conditional) independence relations are implied by the causal structure \cite{Spirtes2000}. In the following examples, we additionally assume \emph{causal sufficiency}, i.\,e.~we assume that there are no hidden confounders.

\subsection{Causal inference in stimulus-based experiments}

Consider two brain state features $X_1$ and $X_2$ in a stimulus-based experiment with $S \not\indep X_1$, $S \indep X_2$, and $S \not\indep X_2 | X_1$.

If we learn an encoding model on this data, we find that $X_1 \in X_S^{\text{enc}}$ and $X_2 \in X_0^{\text{enc}}$. We can thus conclude that $X_1$ is an effect of $S$, i.\,e.~$S \to X_1$ (cf.~section \ref{sec:intA}). We can not, however, determine the causal relation between $X_1$ and $X_2$.

If we learn a decoding model, on the other hand, we find that $X_1,X_2 \in X_S^{\text{dec}}$, i.\,e.~we find both features to be relevant for decoding $S$, as $S \not\indep X_2 | X_1$ and $S \to X_1$.

Under the assumptions of faithfulness and causal sufficiency, the only causal structure that can give rise to these observations is $S \to X_1 \leftarrow X_2$, i.\,e.~$X_2$ is a cause of $X_1$ \cite{Pearl2000}.

By learning both an encoding- and a decoding model on the same data, and comparing relevant features, we have thus 
determined the causal relations between the observed variables. An example of this inference procedure, known as the inference rule for potential causation \cite{Pearl2000}, is given in \cite{Grosse-WentrupNeuroImage2011}.

\subsection{Causal inference in response-based experiments}\label{sec:respcausalinf}

Consider two brain state features $X_1$ and $X_2$ in a response-based experiment with $X_1 \not\indep R$, $X_2 \not\indep R$ and $X_2 \indep R | X_1$.

If we learn an encoding model on this data, we find that $X_1,X_2 \in X_R^{\text{enc}}$ as $X_1,X_2 \not\indep R$. We thus conclude that both $X_1$ and $X_2$ are potential but not necessarily genuine causes of $R$ (cf.~section \ref{sec:antienc}).

If we learn a decoding model, on the other hand, we find that only $X_1 \in X_R^{\text{dec}}$, as $X_2$ does not help for decoding if $X_1$ is already known due to $X_2 \indep R | X_1$. By only looking at the decoding model, we would only identify $X_1$ as a potential cause of $R$.

Taken together, however, the only causal structures that can give rise to these observations, again assuming faithfulness and causal sufficiency, are $X_2 \leftarrow X_1 \to R$ or $X_2 \to X_1 \to R$ \cite{Pearl2000}. As in both structures $X_1 \to R$, we can conclude that $X_1$ is a direct cause of $R$. The role of $X_2$, however, remains ambiguous.

By learning both an encoding- and a decoding model on the same data, and comparing relevant features, we have 
thus again identified a causal relation between observed variables.

\section{Conclusion}\label{sec:conclusion}

In the previous section, we have demonstrated on two examples how the combination of encoding- and decoding models can resolve ambiguities that can not be decided when only looking at one type of model. This is due to the fact that relevant features are determined by univariate independence tests in encoding models and by multivariate conditional independence tests in decoding models. Both types of tests provide complementary information on the underlying causal structure.

As we have shown in section \ref{sec:respcausalinf}, however, these tests do not always uniquely determine the causal structure of a given set of observed variables. In general, conditional independence tests on all subsets of observed variables may provide further information \cite{Pearl2000,Spirtes2000}. An exhaustive description of the causal inference rules based on conditional independence tests is beyond the scope of the present paper.

We conclude by emphasizing that the causal structure, as determined by a priori knowledge and/or causal inference methods, has to be taken into account when interpreting neuroimaging data.
\ \\

\bibliographystyle{IEEEtran}
\bibliography{bibfile}

\end{document}